\documentclass[manuscript,screen]{acmart}
\usepackage{amsmath}

\AtBeginDocument{%
  \providecommand\BibTeX{{%
    \normalfont B\kern-0.5em{\scshape i\kern-0.25em b}\kern-0.8em\TeX}}}

\begin{document}

\title{An Assessment on Comprehending Mental Health through Large Language Models}

\author{Mihael Arcan}
  \email{mihael@luahealth.io}
  \orcid{0000-0002-3116-621X}
\affiliation{%
 \institution{Lua Health}
 \city{Galway}
 \country{Ireland}}

\author{David-Paul Niland}
\email{david-paul@luahealth.io}
\affiliation{%
 \institution{Lua Health}
 \city{Galway}
 \country{Ireland}}


\author{Fionn Delahunty}
\email{fionn@luahealth.io}
\orcid{0000-0002-2185-924X}
\affiliation{%
 \institution{Lua Health}
 \city{Galway}
 \country{Ireland}}

\renewcommand{\shortauthors}{}

\begin{abstract}
Mental health challenges pose considerable global burdens on individuals and communities. Recent data indicates that more than 20\% of adults may encounter at least one mental disorder in their lifetime. On the one hand, the advancements in large language models have facilitated diverse applications, yet a significant research gap persists in understanding and enhancing the potential of large language models within the domain of mental health. On the other hand, across various applications, an outstanding question involves the capacity of large language models to comprehend expressions of human mental health conditions in natural language. This study presents an initial evaluation of large language models in addressing this gap. Due to this, we compare the performance of Llama-2 and ChatGPT with classical Machine as well as Deep learning models. Our results on the DAIC-WOZ dataset show that transformer-based models, like BERT or XLNet, outperform the large language models.
\end{abstract}


\keywords{AI, Large Language Models, Mental Health}


\maketitle

\section{Introduction}

Progress in large language models (LLMs) has enabled a wide range of applications, but there remains a substantial research gap in comprehending and improving LLMs' potential in the realm of mental health. 
Within various applications, an unresolved query pertains to the extent of LLMs' capability to grasp human mental health conditions expressed in natural language. Mental ill-health issues pose a substantial challenge to individuals and communities on a global scale. Recent data indicates that over 20\% of adults are likely to encounter at least one mental disorder during their lifetime, with 5.6\% experiencing severe psychotic disorders that severely impact their functioning. Moreover, depression and anxiety alone contribute to an annual global economic productivity loss of approximately \$1 trillion.

Over the last ten years, there has been an abundance of research in the fields of natural language processing (NLP) and computational social science aimed at identifying mental ill-health issues using online text data, such as social media content \cite{coppersmith-etal-2015-clpsych,GUNTUKU201743}. However, a majority of these investigations have concentrated on constructing specialised machine learning (ML) models for specific tasks like stress detection \cite{Guntuku2018UnderstandingAM} or depression prediction \cite{10.1145/3448107,doi:10.1073/pnas.1802331115}. Even conventional pre-trained language models like BERT \cite{devlin2018pretraining} necessitate fine-tuning for particular downstream tasks. Some studies have explored multi-task setups \cite{benton2017multitask}, like simultaneously predicting depression and anxiety, but they are often limited to predefined task sets, offering minimal flexibility \cite{10.1109/ASONAM55673.2022.10068655}.

On another front, a separate line of research has explored the use of chatbots in mental health services \cite{10.1007/978-3-030-17705-8_11,10.1145/3392836}. Most chatbots operate on rule-based systems and could benefit from more advanced models that enhance their capabilities \cite{ABDALRAZAQ2019103978}. Despite the growing body of research aimed at empowering AI for mental ill-health, it is crucial to recognise that existing techniques may inadvertently introduce biases and even provide harmful advice to users \cite{Chen2019CanAH,0815e82c4cd14a5bb571d88b665d7fcc}.

Anxiety disorders are marked by an excessive sense of fear and worry, accompanied by related behavioral disturbances. These symptoms are severe enough to cause significant distress or impair daily functioning. There are various types of anxiety disorders, including generalised anxiety disorder (characterised by excessive worrying), panic disorder (defined by panic attacks), social anxiety disorder (involving excessive fear and apprehension in social situations), separation anxiety disorder (involving intense fear or anxiety regarding separation from emotionally significant individuals), and others. Effective psychological interventions are available, and in certain cases, depending on factors like age and the severity of the condition, medication may also be considered as a potential treatment option

Depression stands apart from the usual shifts in mood and brief emotional reactions to daily challenges. During a depressive episode, an individual consistently experiences a low mood (such as feeling sad, irritable, or empty) or a diminished capacity to experience pleasure or interest in activities, enduring for the majority of the day, virtually every day, for a minimum of two weeks. Additionally, several other symptoms manifest, which may encompass difficulties in concentration, overwhelming feelings of guilt or diminished self-esteem, a sense of hopelessness about the future, contemplations of death or suicide, disrupted sleep patterns, alterations in appetite or body weight, and heightened fatigue or reduced energy levels. It's crucial to note that individuals with depression face an elevated risk of suicidal thoughts and actions. Fortunately, effective psychological therapies are available, and depending on factors like age and the severity of the condition, medication may also be considered as a viable treatment option.

\section{Related Work}

To address Major Depressive Disorder (MDD), \cite{fionn_aics18} proposes a passive diagnostic system that integrates clinical psychology, machine learning, and conversational dialogue systems. Through the use of sequence-to-sequence neural networks, a real-time dialogue system engages individuals, while specialised machine learning classifiers monitor conversations to predict critical depression symptoms. Evaluation results indicate potential advancements in human-like chatbots and depression identification. Despite acknowledging limitations in data representation and a small sample size, the study suggests the possibility of enhancing support for individuals with MDD through real-time communication tools. Similarly, \cite{delahunty_smm4h19} introduces a deep neural network for predicting PHQ-4 scores (depression and anxiety levels) from written text. Leveraging the Universal Sentence Encoder and a deep learning Transformer neural network, the model demonstrates efficacy in psychometric score prediction. Exploring application to social media data, the study incorporates psycholinguistic features and a multi-dimensional deep neural network, noting challenges related to domain-specificity and generalizability. In addressing MDD, \cite{DBLP:journals/braininf/MilintsevichSD23} employs natural language processing to create a neural classifier detecting depression from speech transcripts. By predicting individual depression symptoms, the study utilises a symptom network analysis approach and achieves comparable results to state-of-the-art models in binary diagnosis and depression severity prediction. Similarly, \cite{DBLP:conf/emnlp/BhatHABL21} focuses on toxic workplace communication in emails, introducing ToxiScope, a taxonomy to detect and quantify toxic language patterns. Through annotation tasks and machine learning models, the study reveals insights into implicit and explicit workplace toxicity. The research suggests refining detection methods and exploring correlations between toxicity, power dynamics, and biases in workplace communication for future directions. \cite{Jiang2023.09.11.23295212} explore the potential of objective digital biomarkers in assessing psychiatric disorders. By investigating behavioral and physiological signals extracted from remote interviews, the research assesses the complementary information provided by multimodal features. The study derives time series features from four conceptual modes: facial expression, vocal expression, linguistic expression, and cardiovascular modulation. These features are extracted from audio and video recordings of remote interviews, using task-specific and foundation models. Four binary classification tasks are defined, including the detection of clinically-diagnosed psychiatric disorders, major depressive disorder, self-rated depression, and self-rated anxiety. Results indicate statistically significant feature differences between controls and subjects with mental ill-health conditions, with correlations found between features and self-rated depression and anxiety scores. The best unimodal performance is achieved by visual heart rate dynamics, with areas under the receiver-operator curve (AUROCs) ranging from 0.68 to 0.75. Combining multiple modalities enhances performance, yielding AUROCs of 0.72 to 0.82. Task-specific models outperform foundation models, suggesting the effectiveness of specific features. This comprehensive multimodal analysis on 73 subjects, using remotely-recorded telehealth interviews, reveals informative characteristics of clinically diagnosed and self-rated mental health status. The study provides early evidence of the utility of multimodal digital biomarkers extracted from low-cost, non-lab-controlled data, offering insights into the most suitable modalities and methods for automated remote mental health assessments.

Large language models (LLMs), such as GPT-3, GPT-4, and Google's PaLM, show potential to revolutionise psychotherapy by supporting, augmenting, or possibly replacing human-led interventions, addressing challenges in mental healthcare capacity and providing personalised treatments. \cite{Stade2023} offer a roadmap for the responsible application of clinical LLMs in psychotherapy. It provides a technical overview, discusses integration stages with parallels to autonomous vehicle technology, explores potential applications in clinical care, training, and research, and offers recommendations for responsible development and evaluation. While recognizing the promise of LLMs, the paper urges caution, emphasizing the need for psychologists to approach integration with care, educate the public about risks, and actively engage with technologists. It advocates for ongoing monitoring and advocacy for responsible and ethical use of LLMs in psychotherapy to ensure patient well-being. \cite{chung2023challenges} delves into primary challenges in LLM development for psychological counseling, addressing model hallucination, interpretability, bias, privacy, and clinical effectiveness. Practical solutions are suggested based on the authors' experiences. While medical applications of LLMs have been experimented with, superior performance over human doctors is noted alongside evidence of limitations in mental health counseling. AI acceptance in medicine, especially mental ill-health, requires substantial improvements and responsible development. The paper anticipates LLM integration into the medical field, acknowledging increased regulatory scrutiny. Challenges are detailed from academic research, advocating a holistic approach and emphasizing diverse data representation and cautious model fine-tuning. Regulatory bodies are expected to play pivotal roles, demanding evidence of beneficial model use. Despite potential shifts in the AI landscape, leveraging the current paradigm is deemed imperative amid escalating global mental health disorders. The paper underscores the necessity of enhancing existing LLMs for psychological counseling tools, recognizing exceptional benefits despite formidable challenges. \cite{doi:10.1212/WNL.0000000000207967} highlight the increasing attention on generative artificial intelligence, particularly large language models (LLMs), and their potential in analyzing extensive medical records for insights into neurology. The paper explores various use cases for LLMs in neurology, including early diagnosis, patient and caregiver support, and assisting clinicians. It acknowledges potential ethical and technical challenges, such as privacy concerns, data security, biases in training data, and the importance of rigorous validation. While recognizing these challenges, the paper emphasises the promising opportunities LLMs present for enhancing the care and treatment of neurologic disorders, underscoring the need for responsible research practices.

\cite{yang2023mentallama} addresses the growing significance of social media as a valuable source for automatic mental health analysis, focusing on interpretable models. While recent large language models (LLMs) have been explored for this purpose, their unsatisfactory performance in zero-shot/few-shot scenarios poses challenges. To tackle this, the authors introduce the first multi-task and multi-source interpretable mental health instruction (IMHI) dataset, containing 105K data samples from diverse social media sources. They use ChatGPT to generate explanations and rigorously evaluate correctness, consistency, and quality. The resulting MentaLLaMA, an open-source instruction-following LLM series, achieves state-of-the-art performance, generating ChatGPT-level explanations and displaying strong generalizability to unseen tasks on the IMHI benchmark. The paper contributes a comprehensive dataset and model for interpretable mental health analysis on social media. \cite{xu2023mentalllm} present a comprehensive evaluation of various Language Models (LLMs), including Alpaca, Alpaca-LoRA, FLAN-T5, GPT-3.5, and GPT-4, in the context of mental ill-health prediction tasks using online text data. The experiments cover zero-shot prompting, few-shot prompting, and instruction finetuning, revealing key insights. The context enhancement strategy consistently improves performance for all LLMs, and mental health enhancement proves effective for models with a substantial number of trainable parameters. Few-shot prompting consistently boosts model performance, even with just one example per class. Crucially, instruction finetuning across multiple datasets significantly enhances model performance across various mental health prediction tasks. The top finetuned models, Mental-Alpaca and Mental-FLAN-T5, outperform larger models like GPT-3.5 and GPT-4, performing on par with the state-of-the-art task-specific model Mental-RoBERTa. An exploratory case study on reasoning capabilities underscores both promising potential and notable limitations of LLMs. The findings are distilled into guidelines for researchers, developers, and practitioners enhancing LLMs' understanding of mental health for downstream tasks. Emphasis is placed on the ethical considerations in this research domain, highlighting that practical deployment of LLMs in mental health applications is currently distant. \cite{Bill1782678} focus on fine-tuning an LLM for a specific function in psychology using Reinforcement Learning from Human Feedback (RLHF) and explores its viability. The theoretical foundation of LLMs, RLHF, and the ethical considerations of developing a psychological AI are presented. Previous studies on RLHF and AI in psychology demonstrate the feasibility of the proposed goal. The methodology for training and evaluating the model involves comparing a pre-trained model with the fine-tuned one, whereby the study finds no clear difference between the used models. The study also delves into an ethical framework for a digital psychology assistant, proposing a suitable introduction to the market and the division of responsibilities. The discussion extends to rules and regulations applicable to this research field, emphasizing the need for governments to introduce relevant regulations for AI innovation, encompassing ethics, accountability, and personal data protection. \cite{galatzerlevy2023capability} investigate the capability of Large Language Models (LLMs), specifically Med-PaLM 2, trained on extensive medical knowledge, to predict psychiatric functioning from patient interviews and clinical descriptions without specific training for such tasks. Analyzing 145 depression, 115 PTSD assessments, and 46 clinical case studies across various disorders, the results indicate that Med-PaLM 2 can assess psychiatric functioning effectively. The strongest performance is observed in predicting depression scores (Accuracy range= 0.80 - 0.84), which is statistically indistinguishable from human clinical raters. The findings suggest the potential of general clinical language models to flexibly predict psychiatric risk based on free descriptions from both patients and clinicians.

\cite{jin2023psyeval} address the growing interest in utilizing LLMs in mental health research and identifies a lack of a comprehensive benchmark for evaluating their capabilities in this domain. The authors introduce PsyEval, the first benchmark tailored to the unique characteristics of mental ill-health, consisting of six sub-tasks across three dimensions. Eight advanced LLMs are evaluated using PsyEval, revealing significant room for improvement in current LLMs in mental health-related tasks. Notably, GPT-4 shows satisfactory performance in mental health Question-Answering (QA) but still requires advancement. The benchmark highlights performance gaps, particularly in tasks involving disease prediction from social media posts and accurate forecasting of depression and suicide severity in simulated doctor-patient dialogues. The results emphasise the need for further advancements in tailoring language models for mental health applications, and PsyEval is positioned as a valuable tool for assessing and guiding such developments.

\section{Methodology}
Within this section we provide our methods on data preprocessing as well as the predictive models used in this work. We further provide insights on leveraging prompts for finetuning the used LLM.

\subsection{Data Preprocessing}

For every participant in the dataset, there is a corresponding PHQ-4 score that is divided into four variables. These variables align to each of the variables in the PHQ-4 questionaire; Generalised Anxiety Disorder 1 and 2 (GAD-1, GAD-2), Patient Health Questionaire 1 and 2 (PHQ-1 and PHQ-2). GAD 1 and GAD 2 relate to anxiety and PHQ-1 and PHQ-2 relate to depression. One participant may have many messages, but that participant has only one overall PHQ4 score. Their score is divided into the four separate categories; GAD-1, GAD-2, PHQ-1 and PHQ-2. As some of the messages were quite short in the dataset, we decided to lengthen the messages to add more contextual information to a PHQ4 score that might be associated with that particular person. We concatenated messages to a maximum of 50 words per observation. We only concatenated messages together that were from the same person. We did not concatenate messages together that were from two different people. After the concatenation step and duplicates and missing values removed, there were 28,186 observations in the training set and 8,710 in the test set. 

\subsection{XGBoost}
The messages were loaded into a pre-trained Transformer model, which produced a dense vector representation of the text. This vector was then loaded into the XGBoost classifier. We used five-fold cross validation to find the optimal hyperparameters. The best model was selected based on weighted F1. 

\subsection{Prompting LLMs}

In order to retrieve a specific response from the LLMs, we prompted the targeted LLMs with five different queries and evaluated each of them on their performance. The variation differed in terms of lexicalised outputs as well as the definitions and length of the promps provided to the LLM. Some initial exploratory experiments led to the formation of templates to structure the prompts that were used. We tailored each prompting strategy to each one of the PHQ or GAD questions. This meant that there was a total of 20 different prompts. 
Differences between versions of prompts only varied slightly. Some differences were minor, such as using "on a scale of zero to three" in one prompt and "one a scale of 0 to 3" in another. Some prompts were brief (cf. Version 3 in \ref{tab:prompting_examples1}), while others were lengthy and included examples of messages that had been evaluated based on specific GAD or PHQ questions within the prompt.

The goal of the prompts was that the LLM would return a number corresponding to the PHQ or GAD question that the prompt was structured around, whereby we had regular expressions to handle the outputs. These consisted of taking the first digit number out of the string or converting numbers in linguistic format to integers. For example, if the chatbot responded with "This messages looks like it is showing zero levels of anxiety", then our system would return "0". This was to make it easier to analyse and evaluate.

\begin{table}[]
    \begin{tabular}{l|p{130mm}}
    \toprule
        Version 1 \\
        \midrule
        GAD-1 & On a scale of \textbf{zero} to \textbf{three}, estimate how anxious this person is with zero being a little anxious and three being incredibly anxious: + \texttt{TEXT}\\
        GAD-2 & On a scale of \textbf{zero} to \textbf{three}, estimate how worried this person is with zero being a little worried and three being incredibly worried: + \texttt{TEXT}\\
        PHQ-1 &  On a scale of \textbf{zero} to \textbf{three}, estimate how depressed this person is with zero being a little depressed and three being incredibly depressed: + \texttt{TEXT}\\
        PHQ-2 & On a scale of \textbf{zero} to \textbf{three}, estimate if this person is expressing no interest in previously enjoyed activities with zero being a little and three being incredibly: + \texttt{TEXT}\\
        \midrule

        Version 2 \\
        \midrule
        GAD-1 & On a scale of \textbf{0} to \textbf{3}, estimate how anxious this person is with 0 being a little anxious and 3 being incredibly anxious: + \texttt{TEXT}\\
        GAD-2 & On a scale of \textbf{0} to \textbf{3}, estimate how worried this person is with 0 being a little worried and 3 being incredibly worried: + \texttt{TEXT}\\
        PHQ-1 & On a scale of \textbf{0} to \textbf{3}, estimate how depressed this person is with 0 being a little depressed and 3 being incredibly depressed: + \texttt{TEXT}\\
        PHQ-2 & On a scale of \textbf{0} to \textbf{3}, estimate if this person is expressing no interest in previously enjoyed activities with 0 being a little and 3 being incredibly: + \texttt{TEXT}\\
        \midrule

        Version 3 \\
        \midrule
        GAD-1 & On a scale of \textbf{zero} to \textbf{three}, rate the anxiety of this message: + \texttt{TEXT}\\
        GAD-2 & On a scale of \textbf{zero} to \textbf{three}, rate the worry in this message: + \texttt{TEXT}\\
        PHQ-1 & On a scale of \textbf{zero} to \textbf{three}, rate the depression in this message: + \texttt{TEXT}\\
        PHQ-2 & On a scale of \textbf{zero} to \textbf{three}, rate the interest in previously enjoyed activities in this message: + \texttt{TEXT}\\
        \midrule
        Version 4 \\
        \midrule
        GAD-1 & \textbf{Generalised anxiety disorder is a mental health illness that is defined by people having feelings of excessive anxiety.} On a scale of zero to three, rate the anxiety in this message: + \texttt{TEXT}\\
        GAD-2 & \textbf{Generalised anxiety disorder is a mental health illness that is defined by people having feelings of excessive worry.} On a scale of zero to three, rate the anxiety in this message: + \texttt{TEXT}\\
        PHQ-1 & \textbf{Depression, or major depressive disorder is a mental health illness that is categorised by people feeling down, depressed or hopeless.} On a scale of zero to three, rate the depression in this message: + \texttt{TEXT}\\
        PHQ-2 & \textbf{Depression, or major depressive disorder is a mental health illness that is categorised by people having little interest or pleasure in doing things.} On a scale of zero to three, rate the depression in this message: + \texttt{TEXT}\\
    \bottomrule
    \end{tabular}
    \caption{Examples for different prompting versions for GAD and PHQ questions.} \vspace{20mm}
    \label{tab:prompting_examples1}
\end{table}

\begin{table}[]
    \centering
    \begin{tabular}{c|p{130mm}}
    \toprule
        Version 5 \\
        \midrule
        GAD-1 & \small Generalised anxiety disorder is a mental health illness that is defined by people having feelings of exessive anxiety. \textbf{The anxiety in this example is rated zero:} "like uh a a a guy that likes to see different sights and go different places uh like different i'm a big movie person so i like going to theatres stuff like that try to free myself from the situation to clear my mind usually take a walk or something" \textbf{The anxiety in this example is rated one:} "got stuck out here then i had my baby and i just stayed be doing a little bit of everything a good night's sleep i grew up not getting to know who my mom was out there just on the food basis they like to give you a lotta food" \textbf{The anxiety in this example is rated two:} "it was it was great and now it's just okay i'm hardworking i'm uh i care about everyone i yep i suppose going on a trip to canada with my daughter was very memorable i love la i love the traffic on the four o five i love the beach" 
        \textbf{The anxiety in this example is rated three:} "regret right now is um that's just how she is value we were always what what's the word i'm looking for um also the weather was just it was absolutely awful there and it was just it was very difficulties in my life she's very touchy i would say in." On a scale of zero to three, rate the anxiety in this message: + \texttt{TEXT} \\ 
        GAD-2 & \small Generalised anxiety disorder is a mental health illness that is defined by people having feelings of excessive worry. \textbf{The worry in this example is rated zero:} "i've added that to my life she's a great woman and uh that started a future in in athletics and that's what i wanted to do so i accomplished my goals so at the time it was a great a great thing and it still is a great thing it's" \textbf{The worry in this example is rated one:} "lot of stress every day just everyday stress uh yeah pretty much that's what triggers it play like games like i like to play video games those are fun stuff like that okay thanks eh my family it's just eh we're not like too too close we kinda argue a" \textbf{The worry in this example is rated three:} "guess that and not and then there's you i don't completely lose it for days at a time when i can get really into that and sort of shut out the rest of the world um the music and the the thoughts just kind of it really makes me a" On a scale of zero to three, rate the anxiety in this message: + \texttt{TEXT}\\
        PHQ-1 & \small Depression, or major depressive disorder is a mental health illness that is categorised by people feeling down, depressed or hopeless. \textbf{The depression in this example is rated zero:} "like uh a a a guy that likes to see different sights and go different places uh like different i'm a big movie person so i like going to theatres stuff like that try to free myself from the situation to clear my mind usually take a walk or something" \textbf{The depression in this example is rated one:} "of my friend i wish i would've handled his sister a little differently as far as the dirt it has been yes it was uh it's very close um no i have not we've always maintained our friendship um i can't recall one off hand no problem and that's all" \textbf{The depression in this example is rated two:} "fight or anything or yeah any of that my grandma she's always giving me encouragement and um she's a therapist a licensed therapist so she's always you know made it really really um yeah i'm okay yeah um i just try to stay positive i try to think like okay" \textbf{The depression in this example is rated three:} "to my son my son and daughter-in-law and my daughter and i went out to have hawaiian food got together it's always fun to get together we got together at my house on christmas i enjoyed just getting together with them i guess the newness wore off that just have" On a scale of zthat’sero to three, rate the depression in this message: + \texttt{TEXT}\\
        PHQ-2 & \small Depression, or major depressive disorder is a mental health illness that is categorised by people having little interest or pleasure in doing things. \textbf{The depression in this example is rated zero:} "like uh a a a guy that likes to see different sights and go different places uh like different i'm a big movie person so i like going to theatres stuff like that try to free myself from the situation to clear my mind usually take a walk or something" \textbf{The depression in this example is rated one:} "of my friend i wish i would've handled his sister a little differently as far as the dirt it has been yes it was uh it's very close um no i have not we've always maintained our friendship um i can't recall one off hand no problem and that's all" \textbf{The depression in this example is rated two:} "um feel uninhibited uninhibited and i was better prepared about three years ago um i was happy that he was safe try not to she has a house i mean a roof over her head to resort to the situation that he was in um because and to have been" \textbf{ The depression in this example is rated three:} "i don't sleep well um well i start to like cry a lot and i start to get really irritable um i argued with my mom and sister yesterday it was just something stupid over like yeah i was like okay well this is the problem and it i just" <</SYS>> On a scale of zero to three, rate the depression in this message: + \texttt{TEXT}\\
    \bottomrule
    \end{tabular}
    \caption{Examples for version 5 prompting for the GAD and PHQ questions.}
    \label{tab:prompting_examplesV5}
\end{table}

\subsection{ChatGPT}
For chatGPT, we used the same prompting structure that worked best for Llama-2, which was Version 3 (see Table~\ref{tab:prompting_examples1}). All other aspects of evaluation and extracting information from the chatbot responses were the same for ChatGPT as it was for Llama-2. 

\section{Experimental Setup}

In this section, we provide insights on the models and the dataset used in our work. We further provide information on the questionnaire and the evaluation metrics used to present the outcomes of our work.

\subsection{Models}

\subsubsection{XGBoost}
\label{subsub:xgboost}

XGBoost (Extreme Gradient Boosting) \cite{Chen:2016:XST:2939672.2939785} is a machine learning algorithm used for both classification and regression tasks. XGBoost is based on the gradient boosting framework, which is an ensemble learning technique. It builds an ensemble of decision trees sequentially, where each tree corrects the errors made by the previous ones. The model uses a customizable objective function that needs to be optimised during training. For regression tasks, the objective is often mean squared error (MSE), while for classification tasks, it can be log loss (binary or multiclass). To prevent overfitting, XGBoost incorporates L1 (Lasso) and L2 (Ridge) regularization techniques into the objective function.

\subsubsection{ChatGPT}
GPT-3 (Generative Pre-trained Transformer) follows the decoder-only Transformer architecture and employs attention mechanisms, allowing it to focus on the most relevant segments of input text, using an extensive context of 2048 tokens and an unprecedented 175 billion parameters. The model exhibited remarkable zero-shot and few-shot learning capabilities across various tasks. The training data for GPT-3 is primarily sourced from a filtered version of Common Crawl, contributing to 60\% of the weighted pre-training dataset, comprising 410 billion byte-pair-encoded tokens. Other data sources include 19 billion tokens from WebText2, 12 billion tokens from Books1, 55 billion tokens from Books2, and 3 billion tokens from Wikipedia. GPT-3 was trained on a vast corpus of text and has demonstrated proficiency in programming languages such as CSS, JSX, and Python, among others.

\subsubsection{LLama}
LLaMA-1 (Large Language Model Meta AI) \cite{touvron2023llama1} is a series of large language models developed by Meta AI. The initial release included models with varying parameter sizes: 7 billion, 13 billion, 33 billion, and 65 billion parameters. LLaMA employs the Transformer architecture with some architectural differences, including the use of SwiGLU activation functions, rotary positional embeddings, and root-mean-squared layer normalization. The foundational models were trained on a vast dataset comprising 1.4 trillion tokens from various publicly available sources. Human annotators were involved in AI alignment by providing prompts and evaluating model outputs, and reinforcement learning from human feedback (RLHF) was employed with a new technique based on Rejection sampling followed by Proximal Policy Optimization (PPO). Additionally, LLaMA aimed to improve multi-turn consistency in dialogues using the "Ghost attention" technique during training to respect system messages throughout conversations. LLaMA-2 \cite{touvron2023llama2} remains mostly consistent with that of LLaMA-1 models, with the notable change being the utilization of 40\% more data for training the foundational models. LLaMA-2 encompasses both foundational models and models specifically fine-tuned for dialogues, referred to as LLaMA-2 Chat. Within this work, we leveraged Llama2 model with 13B parameters.

\subsubsection{Transformer Models}
For a comparison to LLMs, we leverage the Transformer models \cite{NIPS2017_3f5ee243}, which rely on a self-attention mechanism, allowing it to capture contextual dependencies in input sequences efficiently. The model consists of an encoder-decoder structure, with multi-head self-attention layers enabling parallelised processing of input tokens. Positional encoding is used to provide information about the token's position in the sequence. The Transformer's attention mechanism facilitates capturing long-range dependencies, making it highly effective for tasks requiring context understanding. Within this work, we compare the BERT and Roberta Transformer models and their distilled versions, i.e. DistilBert and Dsitil-Roberta. In addition to that we leverage the XLNet models as well. The BERT model \cite{devlin2019bert} is pre-trained on large corpora and can then be fine-tuned for specific natural language processing (NLP) tasks, such as text classification, named entity recognition, and question answering, among others. BERT embeddings have been widely adopted and have significantly improved the state-of-the-art performance in various NLP applications. DistilBERT \cite{sanh2020distilbert} is a distilled version of BERT, offering a more compact and faster alternative for tasks in natural language processing (NLP). Despite having fewer parameters, DistilBERT embeddings can be utilised in various NLP applications, providing a balance between computational efficiency and model performance. RoBERTa \cite{liu2019roberta}, or Robustly optimised BERT approach, is a variant of the BERT (Bidirectional Encoder Representations from Transformers) model, uses dynamic masking during pre-training, removing the Next Sentence Prediction (NSP) objective, and training with larger mini-batches and learning rates. These modifications result in a more robust and efficient model. XLNet \cite{10.5555/3454287.3454804} is a Transformer-based language model, which combines ideas from autoregressive language modeling (as seen in models like GPT) and autoencoding (as in BERT) to capture bidirectional context and maintain long-term dependencies in sequences. Instead of predicting the next word in a sentence, XLNet is trained to predict a permutation of the words. This approach allows the model to consider bidirectional context while preventing it from seeing the entire context during training, enhancing its ability to capture dependencies.

For all models, we leverage the base version of the transformer models.

\subsection{DAIC-WOZ Dataset}
Within this comparison, we leveraged the DAIC-WOZ dataset \cite{gratch-etal-2014-distress}, which comprises clinical interviews aimed at aiding the assessment of psychological distress conditions like anxiety, depression, and post-traumatic stress disorder. These interviews were gathered as part of a broader initiative to develop a computer-based system that conducts interviews with individuals and recognises verbal and nonverbal cues associated with mental health issues. Specifically, it encompasses data from Wizard-of-Oz interviews, where an animated virtual interviewer named Ellie, under the control of a human interviewer in a separate location, conducted the interviews. The data, which consists of 189 interaction sessions, each lasting between 7 to 33 minutes, has been originally transcribed and annotated to encompass a range of verbal and non-verbal characteristics.

\subsection{PHQ-4 Questionnaire}
The Patient Health Questionnaire-4 (PHQ-4) \cite{KroenkePhq4} was developed to address the challenge posed by the high prevalence of anxiety and depression in the general population. Since these two mood disorders often co-occur and individuals with these conditions may struggle with fatigue or difficulty concentrating, the PHQ-4 offers a concise and accurate assessment tool. The PHQ-4 consists of four questions, each answered on a four-point Likert-type scale. It serves the purpose of providing a very brief yet precise measurement of the fundamental symptoms associated with depression and anxiety. It combines a two-item measure for depression (PHQ–2), which focuses on core depressive criteria, and a two-item measure for anxiety (GAD–2), both of which have been independently proven to be effective screening tools. The total PHQ–4 score complements the scores of these subscales, offering an overall assessment of symptom burden, functional impairment, and disability. While an elevated PHQ–4 score is not diagnostic, it serves as an indicator for further evaluation to confirm the presence or absence of a clinical disorder that requires treatment.


\subsection{Evaluation Metrics}

Besides analysing the widely used metrics, i.e., \textbf{weighted precision}, \textbf{recall} and \textbf{F1} for our experiments, we extend our metrics with further metrics in the field of statistics and medicine. \textbf{Weighted specificity} describes the accuracy of a test that reports the presence or absence of a medical condition. It can be useful for "ruling in" disease since the test rarely gives positive results in healthy patients. A test with a specificity of 1.0 will recognise all patients without the disease by testing negative, therefore a positive test result would rule in the presence of the disease. Nevertheless, a negative result from a test with high specificity is not necessarily useful for "ruling out" a disease. As an example, a test that always returns a negative test result will have a specificity of 1.0 because specificity does not consider false negatives. A test like that would return negative for patients with the disease, making it useless for "ruling out" the disease.


Furthemore, we leverage the Hamming loss and the AUC-ROC Curve.  
The \textbf{Hamming loss} is a metric used in multi-label classification to quantify the accuracy of predictions by measuring the fraction of incorrectly predicted labels across all instances. It is calculated as the average fraction of incorrectly predicted labels per instance, with a score of 0 indicating perfect predictions and 1 indicating complete misclassification. The Hamming loss accounts for both false positives and false negatives in the predicted label sets, making it a valuable measure for evaluating the overall performance of multi-label classification models. The \textbf{AUC-ROC} (Area Under the Receiver Operating Characteristic Curve) is a graphical representation of a binary classification model's performance across various threshold settings. It plots the true positive rate against the false positive rate, illustrating the trade-off between sensitivity and specificity. The AUC-ROC value quantifies the model's ability to distinguish between classes, with a higher AUC indicating better overall performance.

\section{Results}
Within this section, we provide the insights on evaluating different prompting strategies, as well as how Llama-2 and ChatGPT persorm compared to XGBoost and different Transformer models.

\subsection{Llama-2 Prompting}
Leveraging Llama-2, we prompted the LLM with different lexical inputs as seen in Tables \ref{tab:prompting_examples1} and \ref{tab:prompting_examplesV5}. As seen in Table \ref{tab:prompting}, we evaluated each GAD and PHQ question separately. For GAD-1, prompting Version 3, i.e., \textit{On a scale of \textbf{zero} to \textbf{three}, rate the anxiety of this message}, showed the best performance in terms of weighted F1, as well in weighted precision, recall, Hamming loss and AUC-ROC. For the specificity metric, version 1 slightly outperforms all other prompting options. For GAD-2, the best F1 score is obtained by various prompting versions, with the highest F1 score of 0.53. The best specificity for GAD-2 is achieved by version 3 while prompting with version 1 provides the best AUC-ROC result. In terms of PHQ-1, the best weighted F1 score is obtained using prompting version 1, while the best precision is obtained using version 3 prompting. Similar to the GAD questions, leveraging prompting version 3 for the PHQ-2 question provided the best results in terms of F1, as well as precision and specificity. 

\begin{table}[]
    \setlength{\tabcolsep}{4pt}
    \begin{tabular}{rcccccc|cccccc}
    \toprule
        & \multicolumn{6}{c}{GAD-1} & \multicolumn{6}{c}{GAD-2}  \\
        \midrule
        & Prec & Rec & F1 & Spec & HammL & AUC-ROC & Prec & Rec & F1 & Spec & HammL & AUC-ROC \\
        \cmidrule{2-13}
        Version 1 & 0.31 & 0.29 & 0.22 & \textbf{0.69} & 0.71 & 0.49 & 0.53 & 0.52 & 0.50 & 0.48 & 0.48 & \textbf{0.51} \\
        Version 2 & 0.33 & 0.32 & 0.21 & 0.66 & 0.68 & 0.50 & 0.44 & 0.66 & \textbf{0.53} & 0.33 & 0.34 & 0.50 \\
        Version 3 & \textbf{0.38} & \textbf{0.33} & \textbf{0.33} & 0.68 & \textbf{0.67} & \textbf{0.52} & \textbf{0.56} & 0.23 & 0.27 & \textbf{0.78} & 0.77 & 0.50 \\
        Version 4 & 0.11 & \textbf{0.33} & 0.16 & 0.67 & \textbf{0.67} & 0.50 & 0.44 & \textbf{0.67} & \textbf{0.53} & 0.33 & \textbf{0.33} & 0.50 \\
        Version 5 & 0.11 & \textbf{0.33} & 0.16 & 0.67 & \textbf{0.67} & 0.50 & 0.44 & \textbf{0.67} & \textbf{0.53} & 0.33 & \textbf{0.33} & 0.50 \\
    \midrule
        & \multicolumn{6}{c}{PHQ-1} & \multicolumn{6}{c}{PHQ-2}  \\
        \midrule
        & Prec & Rec & F1 & Spec & HammL & AUC-ROC & Prec & Rec & F1 & Spec & HammL & AUC-ROC \\
        \cmidrule{2-13}
        Version 1 & 0.37 & 0.36 & \textbf{0.32} & 0.64 & 0.64 & 0.50 & 0.29 & 0.35 & 0.26 & 0.64 & 0.65 & \textbf{0.50} \\
        Version 2 & 0.21 & \textbf{0.46} & 0.29 & 0.54 & 0.55 & 0.50 & 0.17 & \textbf{0.41} & 0.24 & 0.59 & \textbf{0.59} & \textbf{0.50} \\
        Version 3 & \textbf{0.43} & 0.29 & 0.30 & \textbf{0.76} & 0.71 & \textbf{0.53} & \textbf{0.34} & 0.34 & \textbf{0.32} & \textbf{0.65} & 0.66 & 0.49 \\
        Version 4 & 0.21 & \textbf{0.46} & 0.29 & 0.54 & \textbf{0.54} & 0.50 & 0.17 & \textbf{0.41} & 0.24 & 0.59 & \textbf{0.59} & \textbf{0.50} \\
        Version 5 & 0.21 & \textbf{0.46} & 0.29 & 0.54 & \textbf{0.54} & 0.50 & 0.17 & \textbf{0.41} & 0.24 & 0.59 & \textbf{0.59} & \textbf{0.50} \\
    \bottomrule
    \end{tabular}
    \caption{Insights on weighted precision, recall, F1, Hamming loss (HammL) and AUC-ROC (Area Under the Receiver Operating Characteristic Curve for different prompting variant (bold scores represent best result for each metric).}
    \label{tab:prompting}
\end{table}

\subsection{Transformer Models}
As a comparison to LLMs, we deployed different baseline transformer models, which were fine-tuned on the DAIC-WOZ dataset. Table \ref{tab:transformers} shows the results for the different GAD and PHQ questions, whereby Distil-Roberta performs best for GAD-1 and GAD-2 in terms of F1. Specificity scores were best using BERT, RoBERTa or XLNet for GAD-1, while for GAD-2, specificity was best using DistilBERT or XLNet.

\begin{table}[]
    \setlength{\tabcolsep}{4pt}
    \begin{tabular}{rcccccc|cccccc}
    \toprule
        & \multicolumn{6}{c}{GAD-1} & \multicolumn{6}{c}{GAD-2}  \\
        \midrule
        & Prec & Rec & F1 & Spec & HammL & AUC-ROC & Prec & Rec & F1 & Spec & HammL & AUC-ROC \\
        \cmidrule{2-13}
        BERT            & 0.56 & 0.56 & 0.53 & \textbf{0.44} & \textbf{0.59} & 0.66 & 0.67 & 0.69 & 0.67 & 0.31 & 0.62 & \textbf{0.52} \\
        DistilBERT      & \textbf{0.59} & \textbf{0.58} & \textbf{0.56} & 0.42 & 0.63 & \textbf{0.72} & 0.65 & 0.67 & 0.65 & \textbf{0.33} & 0.58 & \textbf{0.52} \\ 
        Distil-RoBERTa  & 0.57 & \textbf{0.58} & \textbf{0.56} & 0.42 & 0.63 & 0.70 & \textbf{0.69} & \textbf{0.70} & \textbf{0.68} & 0.30 & 0.62 & \textbf{0.52} \\ 
        RoBERTa         & 0.55 & 0.56 & 0.55 & \textbf{0.44} & 0.62 & 0.70 & 0.67 & \textbf{0.70} & 0.65 & 0.30 & \textbf{0.56} & 0.48 \\
        XLNet           & 0.55 & 0.56 & 0.54 & \textbf{0.44} & 0.61 & 0.70 & 0.64 & 0.67 & 0.62 & \textbf{0.33} & \textbf{0.56} & 0.45 \\
    \midrule
        & \multicolumn{6}{c}{PHQ-1} & \multicolumn{6}{c}{PHQ-2}  \\
        \midrule
        & Prec & Rec & F1 & Spec & HammL & AUC-ROC & Prec & Rec & F1 & Spec & HammL & AUC-ROC \\
        \cmidrule{2-13}
        BERT            & 0.50 & 0.52 & 0.49 & \textbf{0.48} & \textbf{0.59} & 0.68 & 0.53 & 0.54 & 0.52 & \textbf{0.46} & \textbf{0.61} & 0.71 \\
        DistilBERT      & 0.53 & 0.54 & 0.52 & 0.46 & 0.60 & 0.69 & 0.57 & 0.57 & 0.54 & 0.43 & \textbf{0.61} & 0.72\\ 
        Distil-RoBERTa  & 0.55 & 0.55 & 0.53 & 0.45 & 0.61 & \textbf{0.71} & 0.55 & 0.57 & 0.54 & 0.43 & 0.62 & \textbf{0.73} \\ 
        RoBERTa         & 0.56 & 0.56 & 0.54 & 0.44 & 0.61 & 0.69 & 0.57 & 0.57 & \textbf{0.56} & 0.43 & 0.64 & \textbf{0.73} \\
        XLNet           & \textbf{0.58} & \textbf{0.58} & \textbf{0.55} & 0.42 & 0.64 & 0.70 & \textbf{0.58} & \textbf{0.58} & \textbf{0.56} & 0.42 & 0.63 & \textbf{0.73} \\
    \bottomrule
    \end{tabular}
    \caption{Insights on weighted precision, recall, F1, Hamming loss (HammL) and AUC-ROC (Area Under the Receiver Operating Characteristic Curve for different Transformer models (bold scores represent best result for each metric).}
    \label{tab:transformers}
\end{table}

\subsection{Overall Comparison}
We further summarise all best approaches obtained by prompting Llama-2, i.e. prompting version 3, and Distil-RoBERTa, which performed overall best on the GAD and PHQ questions. In addition to that, we evaluate the predictions using XGBoost (see Section \ref{subsub:xgboost}) as well as ChatGPT, a commercial LLM built by OpenAI. As seen in Table \ref{tab:overall}, XGBoost always outperforms all used models in terms of the Hamming loss metric. Comparing Llama-2 with ChatGPT, we observe minor advantages using ChatGPT, which outperforms Llama-2 on the GAD-2, PHQ-1 and PHQ-2 questions. Finally, we compare the Distil-RoBERTa transformer model with Llama-2 and ChatGPT. Our study shows that the later outperforms all targeted models for all GAD and PHQ questions in terms of weighted precision, recall and F1 score.

\begin{table}[]
    \setlength{\tabcolsep}{4pt}
    \begin{tabular}{rcccccc|cccccc}
    \toprule
        & \multicolumn{6}{c}{GAD-1} & \multicolumn{6}{c}{GAD-2}  \\
        \midrule
        & Prec & Rec & F1 & Spec & HammL & AUC-ROC & Prec & Rec & F1 & Spec & HammL & AUC-ROC \\
        \cmidrule{2-13}
         XGBoost &  0.45 & 0.55 & 0.48 & 0.64 & \textbf{0.45} & 0.56 & 0.63 & 0.69 & 0.60 & 0.34 & \textbf{0.31} & 0.51\\
         Llama-2 (v3) & 0.38 & 0.33 & 0.33 & 0.68 & 0.67 & 0.52 & 0.56 & 0.23 & 0.27 & \textbf{0.78} & 0.77 & 0.50 \\
         ChatGPT 3.5 & 0.36 & 0.29 & 0.31 &\textbf{ 0.71} & 0.71 & 0.51 & 0.54 & 0.37 & 0.44 & 0.62 & 0.63 & \textbf{0.53} \\
         Distil-RoBERTa & \textbf{0.57} & \textbf{0.58} & \textbf{0.56} & 0.42 & 0.63 & \textbf{0.70} & \textbf{0.69} & \textbf{0.70} & \textbf{0.68} & 0.30 & 0.62 & 0.52 \\ 
    \midrule
        & \multicolumn{6}{c}{PHQ-1} & \multicolumn{6}{c}{PHQ-2}  \\
        \midrule
        & Prec & Rec & F1 & Spec & HammL & AUC-ROC & Prec & Rec & F1 & Spec & HammL & AUC-ROC \\
        \cmidrule{2-13}
         XGBoost & 0.51 & 0.54 & 0.48 & 0.61 & \textbf{0.46} & 0.55 & 0.52 & 0.53 & 0.48 & 0.66 & \textbf{0.47} & 0.56 \\
         Llama-2 (v3) & 0.43 & 0.29 & 0.30 & \textbf{0.76} & 0.71 & 0.53 & 0.34 & 0.34 & 0.32 & 0.65 & 0.66 & 0.49 \\
         ChatGPT 3.5 & 0.38 & 0.39 & 0.39 & 0.64 & 0.61 & 0.52 & 0.37 & 0.31 & 0.33 & \textbf{0.71} & 0.69 & 0.50 \\
         Distil-RoBERTa & \textbf{0.55} & \textbf{0.55} & \textbf{0.53} & 0.45 & 0.61 & \textbf{0.71} & \textbf{0.55} & \textbf{0.57} & \textbf{0.54} & 0.43 & 0.62 & \textbf{0.73} \\ 
    \bottomrule
    \end{tabular}
    \caption{Comparison on weighted precision, recall, F1, Hamming loss (HammL) and AUC-ROC (Area Under the Receiver Operating Characteristic Curve for XGBoost, Llama-2, ChatGPT and Distil-Roberta (bold scores represent best result for each metric).}
    \label{tab:overall}
\end{table}

\section{Conclusions}

In conclusion, mental ill-health challenges globally impact a significant portion of the population, highlighting the pressing need for effective interventions. Despite the advancements of large language models in various NLP tasks and their diverse applications, a substantial research gap persists regarding their understanding and optimisation within the realm of mental health. This study addresses this gap by conducting an initial evaluation of large language models, comparing the performance of Llama-2 and ChatGPT with the classical machine and deep learning models. Leveraging Llama-2 and ChatGPT, we explore different prompting strategies and evaluate their effectiveness on GAD and PHQ questions. The results indicate that transformer-based models, such as BERT or XLNet, outperform large language models with a significantly larger parameter set. Notably, Distil-RoBERTa consistently outperforms all models for all GAD and PHQ questions in terms of weighted precision, recall, and F1 score. These findings contribute valuable insights for the future development and application of language models in addressing mental health concerns. Nevertheless the outcomes of our initial study, we will further study large language models and how to adapt them to the challenges in mental ill-health. Due to the sensitive nature of mental health information, we will further analyse biases in training data and the dynamic nature of mental health that are the biggest hurdles in achieving comprehensive and unbiased model performance.

\bibliographystyle{ACM-Reference-Format}
\bibliography{sample-base}










\end{document}